\documentclass{mva_style_arxiv}
\usepackage{graphicx}
\usepackage{pifont}
\usepackage{amsmath}
\usepackage{array, multirow}
\usepackage{color}

\finalcopy %Uncomment this line for the Camera-Ready Manuscript

\begin{document}
% \title{Multi-Modal Faster R-CNN for Pedestrian Detection Considering Misalignment}
%%%%% TSBThttps://ja.overleaf.com/project/604a18a647c07f33ca43c550
%\title{Modal-Wise Regression with Multi-Modal IoU: A Simple Baseline for Multi-Modal Pedestrian Detection with Large Misalignment }
% \title{Modal-wise Regression and Multi-modal IoU: \\ Multi-modal Pedestrian Detection with Large Misalignment }
\title{Multi-Modal Pedestrian Detection with Large Misalignment\\ Based on Modal-Wise Regression and Multi-Modal IoU}

\author{
  Napat Wanchaitanawong\textsuperscript{1,*}, Masayuki Tanaka\textsuperscript{2,*}, Takashi Shibata\textsuperscript{3,**}, Masatoshi Okutomi\textsuperscript{4,*}\\
  Tokyo Institute of Technology\textsuperscript{*}, NTT Corporation\textsuperscript{**}\\
  Tokyo 152-8550, Japan\textsuperscript{*}, Kanagawa 243-0198, Japan\textsuperscript{**}\\
  %Department of Systems and Control Engineering, School of Engineering,\\
  %Tokyo Institute of Technology, Meguto-ku, Tokyo 152-8550, Japan\\
  {\fontsize{11}{13}\selectfont\tt wnapat@ok.sc.e.titech.ac.jp\textsuperscript{1},\{mtanaka\textsuperscript{2},mxo\textsuperscript{4}\}@sc.e.titech.ac.jp,t.shibata@ieee.org\textsuperscript{3}}
}

\maketitle

% \footnotetext[1]{Department of Systems and Control Engineering, School of Engineering, Tokyo Institute of Technology, Meguto-ku, Tokyo 152-8550, Japan}
% \footnotetext[2]{Otemachi First Square, East Tower, 5-1, Otemachi 1-Chome, Chiyoda-ku, Tokyo 100-8116, Japan}

\section*{\centering Abstract}
\textit{The combined use of multiple modalities enables accurate pedestrian detection under poor lighting conditions by using the high visibility areas from these modalities together. The vital assumption for the combination use is that there is no or only a weak misalignment between the two modalities. In general, however, this assumption often breaks in actual situations. Due to this assumption's breakdown, the position of the bounding boxes does not match between the two modalities, resulting in a significant decrease in detection accuracy, especially in regions where the amount of misalignment is large. In this paper, we propose a multi-modal Faster-RCNN that is robust against large misalignment. The keys are 1) modal-wise regression and 2) multi-modal IoU for mini-batch sampling. To deal with large misalignment, we perform bounding box regression for both the RPN and detection-head with both modalities. We also propose a new sampling strategy called ``multi-modal mini-batch sampling" that integrates the IoU for both modalities. 
%To the best of our knowledge, this paper is the first work to tackle the problem "object detection from multi-modal images with large misalignment". 
We demonstrate that the proposed method's performance is much better than that of the state-of-the-art methods for data with large misalignment through actual image experiments.
}

\section{Introduction}
\begin{figure}[t]
  \begin{center}
    \includegraphics[width=60mm]{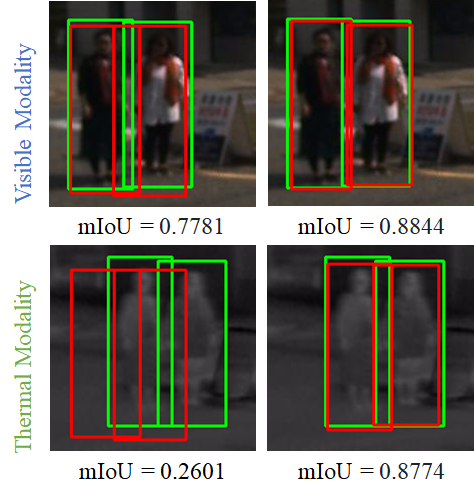}
    \put(-168,-6){(a) Existing method}
    \put(-74,-6){(b) Proposed method}
  \end{center}
  \caption{Visualization examples of ground truth annotations by~\cite{arcnn} (boxes in green), detection results (boxes in red), and mIoU between them. Image patches are cropped from visible-thermal image pairs on the same position.}
\vspace{-6mm}
\end{figure}
Pedestrian detection is still an important issue in machine vision and its applications. 
In practical situations, detection accuracy is significantly degraded under poor lighting conditions when only visible images are used~\cite{ccf,fasterforpedestrian,whathelp,scaleaware,occlusionaware}.
To achieve a robust pedestrian detection in the poor lightning condition, various approaches have been proposed to combine multiple modalities (e.g., visible and far-infrared)~\cite{kaist,pedestriandaynight}. %and to use the highly visible regions by these modalities together.
%The combination use's critical assumption
The critical assumption for the fusion
is that there is no or only weakly misalignment between the two modalities.
In general, however, these assumptions often break down due to lack of time synchronization, inaccurate calibration, or the effects of disparity for stereo~\cite{cats,kaistdriving}.

Recently, to address this issue, several methods that are robust to weak misalignment have been proposed.
For example, L. Zhang et al.~\cite{arcnn} proposed incorporating a module inside the Faster-RCNN that predicts and then aligns the travel distance between modalities for each region.
In general, however, this existing method assumes weak misalignment and is very sensitive to large misalignment.
As shown in Fig.~1~(a),  the position of bounding boxes detected by the existing algorithms are identical in both modalities, resulting in the poor mean intersection over union (mIoU) in thermal images, both pedestrians in thermal modality would be evaluated as false negatives. Thus, the existing methods often fail to detect in either (or both) modal when the misalignment is large. 
In summary, multi-modal image detection with large misalignment is an unsolved problem, even though it is typical for machine vision with multi-modal information.

In this paper, we propose a multi-modal Faster-RCNN that is robust against large misalignment.
To the best of our knowledge, this paper is the first work to address the problem of ``object detection from multi-modal images with large misalignment".
The keys are 1) a new sampling strategy called ``mini-batch sampling based on the amount of misalignment" by introducing a new metric called multi-modal IoU, and 2) modal-wise regression: bounding-box regression for each modal to deal with large misalignment. 
Using the proposed method, the correct bounding boxes detect objects in both visible and thermal images with high mIoU, as shown in Fig.~1~(b).
Real image experiments show that the proposed method's performance significantly outperforms that of state-of-the-art methods for data containing large misalignment.

This paper's contributions are as follows: 1) a new problem: the detection from multi-modal images with large misalignment, 2) modal-wise regression to deal with the large misalignment, and 3) multi-modal IoU and mini-batch sampling strategy for training for multi-modal inputs.

\begin{figure}[t]
  \begin{center}
    \includegraphics[width=75mm]{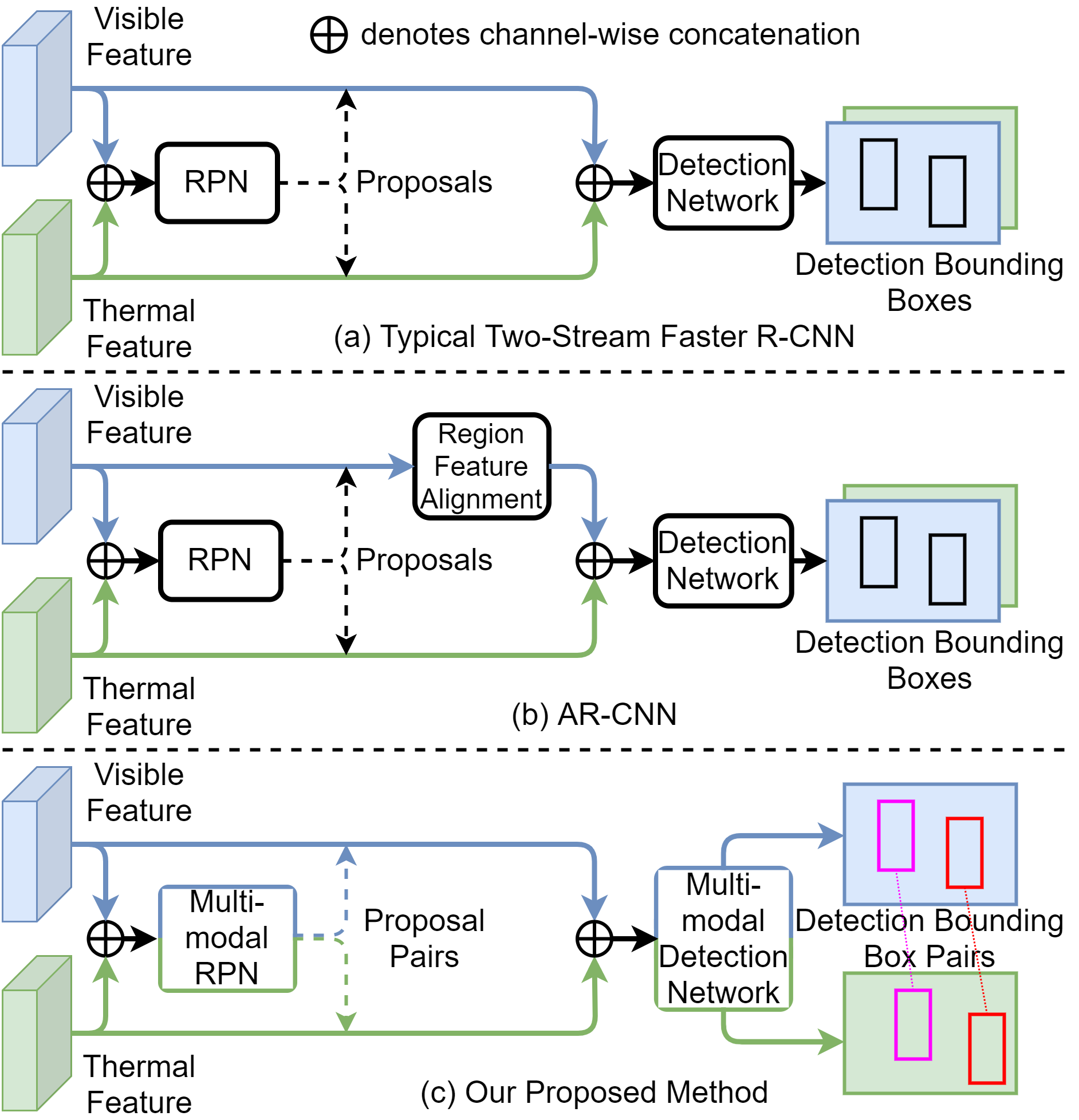}
  \end{center}
  \vspace{-4mm}
  \caption{{\bf{Comparison between multi-modal pedestrian detection frameworks.}} Our proposed method produces pairs of bounding boxes for both modalities as output. Blue and green blocks/paths represent properties of visible and thermal modality respectively.}
\vspace{-6mm}
\end{figure}

\section{Related Work}

\noindent
\textbf{Multi-modal pedestrian detection.} 
%KAIST dataset \cite{kaist} assists this research field to progress steadily. 
KAIST Multispectral Pedestrian Detection (KAIST) dataset~\cite{kaist} has been widely used in the research field of multi-modal pedestrian detection.
Despite non-CNN-based approach such as Aggregate Channel Features (ACF)~\cite{acf} in the early days, the CNN-based approach is mainstream in this field currently~\cite{liu,Wagner2016MultispectralPD,accum,rpnmulti,learningcrossmodal,unified,msds,fusionseg,liu2019illumination,zhang2019cross,guide}.
The main challenge in the early days was how to combine and make use of information from both modalities as with other computer vision applications~\cite{li2013image,10.1117/1.JEI.28.3.033028,shibata2017misalignment}. 
Most importantly, most of the existing methods strictly assume that visible-thermal image pairs are geometrically aligned. Those methods merely fuse both modalities' features in corresponding pixel position directly, as shown in Fig.~2~(a). 
% Shibata 20210605: add the following sentence for #R4's comment: 
Although many geometric calibration and image alignment methods for multi-modal cameras have been proposed~\cite{ogino2017coaxial, kim2015dasc,dong2018learning}, accurate and dense alignment for each pixel is still open problem.
As a result, their detectors suffer dramatically worse performance in poorly aligned regions.

\vspace{0.2cm}
\noindent
\textbf{Weak misalignment.}
AR-CNN~\cite{arcnn} is the first work that immensely tackles the misalignment issue in multi-modal CNN-based pedestrian detection.
% They analyzed the position shift problem, proposed aligned region CNN (AR-CNN), and provided KAIST-Paired annotation.
They also provided a novel KAIST-Paired annotation.
Their method predicts shift distance between modalities for each Region of Interest (RoI), relocates visible region into the thermal area, then proceeds to align them together, as illustrated in Fig.~2~(b). 
% They successfully improved performance from previous methods that do not consider misalignment, revealing the misalignment's significance. 
% In addition to the weak misalignment, MBNet~\cite{mbnet} also proposes a method that takes Modality Imbalance into account.
MBNet~\cite{mbnet} also proposes a method that takes Modality Imbalance into account.
However, those methods assume that the misalignment is weak, which leads to inaccurate detection of bounding boxes in one (or both) modality in large misalignment.
To tackle this problem, we introduce the modal-wise regressor to detect each object in a pair of bounding boxes with different coordinates in each modality, as shown in Fig.~2~(c), resulting in more accurate object localization in both modalities.

\begin{figure*}[t]
  \begin{center}
    \includegraphics[width=170mm]{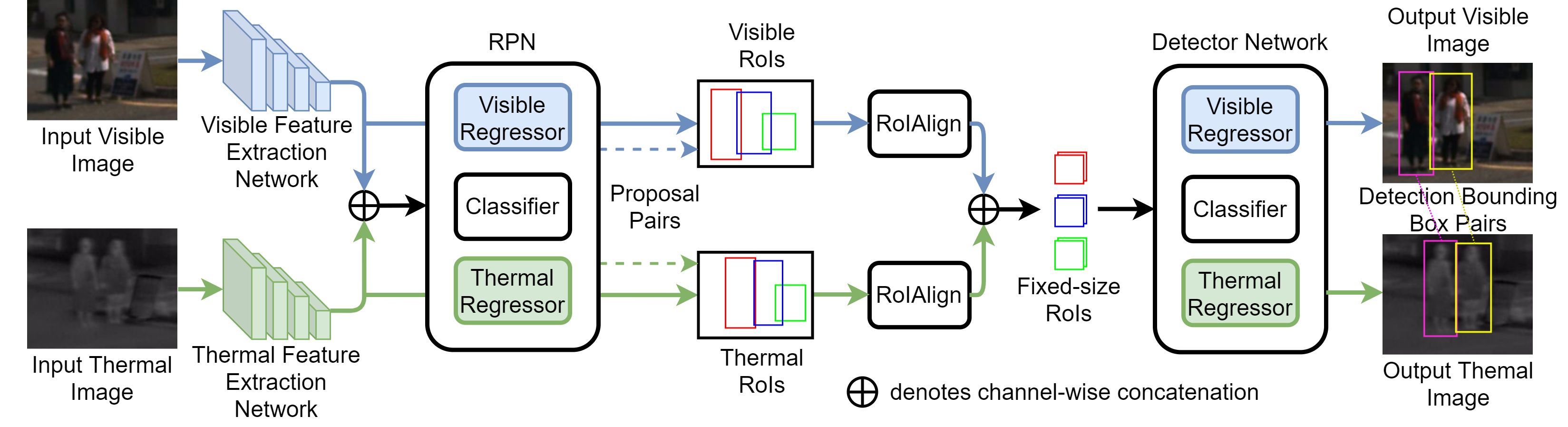}
  \end{center}
  \vspace{-4mm}
  \caption{{\bf{The overall architecture of our network.}} We extend Faster R-CNN into two-stream network to take visible-thermal image pairs as input, then return pairs of detection bounding boxes as output for both modalities. Blue and green blocks/paths represent properties of visible and thermal modality respectively. RoIs and bounding boxes with the same color represent their paired relations.}
\vspace{-6mm}
\end{figure*}  

%\section{Proposed method: modal-wise regression and multi-modal IoU }
\section{Proposed Method}
% To tackle this problem, we introduce the modal-wise regressor to detect each object in a pair of bounding boxes with different coordinates in each modality, as shown in Fig.~2~(c), resulting in more accurate object localization in both modalities.
%%%% Shibata 20210605: remove an "information of" from the following text
We adopt Faster R-CNN~\cite{faster} architecture and extend it into two-stream network for multi-modal imaging, which consists of multi-modal RPN and multi-modal detector. Overview of our network structure is shown in Fig.~3. Moreover, multi-modal IoU (IoU\textsuperscript{M}) and our mini-batch sampling strategy are introduced.

\subsection{Multi-modal RPN}
%We employ RPN into multi-modal RPN by assigning one regressor for each modality exclusively.
The proposed multi-modal RPN has a regressor for each modality.
This will enable proposals from each modality to adjust their sizes and positions independently.
After receiving channel-wise concatenated features from backbone networks, the proposed multi-modal RPN will generate proposal pairs as its output, via classifier predicting each proposal pair a confidence score.    
To keep paired relations of proposals after applying NMS, we use thermal modality proposals as a reference, if any of them are suspended, they also suspend their corresponding pairs in visible modality. 
All remaining proposals will be applied with RoIAlign~\cite{maskrcnn} operation before returning to channel-wise concatenate with their corresponding pairs, resulting in well-aligned RoI for the detector. We employ the loss function of RPN from~\cite{faster} and add one more regression loss to optimize precision of both modals, which is defined as:
\begin{align}
& L(\{p_i\},\{t_i^{V}\},\{t_i^{T}\}) = \frac{1}{N_{cls}}\sum_iL_{cls}(p_i,p_i^*)\\
& + \lambda\frac{1}{N_{reg}}\left[\sum_ip_i^*L^V_{reg}(t_i^{V},t_i^{V*}) + \sum_ip_i^*L^T_{reg}(t_i^{T},t_i^{T*})\right],\nonumber
\end{align}
where i is the index of the anchor, $p_i$ is the predicted probability of anchor i being an object. $p^*_i$ is ground truth label, which equals 1 if anchor i is positive, and equals 0 if anchor i is negative. $t_i^{V}$, $t_i^{T}$ are vector representing coordinates of predicted bounding box pair in visible and thermal modalities respectively. $t_i^{V*}$, $t_i^{T*}$ are ground truth bounding box pair that associate with anchor i. $L_{cls}$ is a cross entropy over object and not object classes. Regression losses $L^V_{reg}$, $L^T_{reg}$ are smooth $L_1$ loss defined in~\cite{fast} for visible and thermal modality respectively. $N_{cls}$ is mini-batch size and $N_{cls}$ is number of anchor locations. We set $\lambda=1$ for all experiments.

\subsection{Multi-modal detector}
Similar to RPN, 
%we adapt detector into multi-modal by assigning one regressor for each modality to adjust bounding boxes' positions in each modality independently. 
the proposed multi-modal detector network has one regressor for each modality to adjust bounding boxes' positions independently, and one classifier to predict each bounding box pair a confidence score.
NMS also works the same way as RPN's. In the end, we will have detection result as pairs of bounding boxes for both modalities, which have different sizes and positions in different modalities, results in detection bounding boxes that are precise for both modalities and also keep their paired relations. We adopt loss function of detector from~\cite{fast} and add one more regression loss, which is defined as:
\begin{align}
& L(p,t^V,t^T) = L_{cls}(p,u)\nonumber\\
& + \lambda|u|\left[L^V_{loc}(t^V,v^V)+L^T_{loc}(t^T,v^T)\right],
\end{align}
where $L_{cls}$ is a cross entropy for class probability $p$ and true class $u$. Regression losses $L^V_{loc}$, $L^T_{loc}$ are smooth $L_1$ loss over predicted regression offsets $t^V$, $t^T$ and regression targets $v^V$, $v^T$ for visible and thermal modality respectively.
$|u|$ is one-hot encoding vector, equals 1 when u is in object classes and 0 otherwise. We set $\lambda=1$ for all experiments.

\subsection{Multi-modal IoU}
Traditionally, we use Intersection-over-Union (IoU) to classify prediction results into true/false positives and negatives categories in evaluation, defined as:
\begin{align}
IoU = \frac{GT\cap DT}{GT\cup DT},
\end{align}
where $GT$, $DT$ denote ground truth bounding boxes and detection bounding boxes respectively. $GT\cap DT$ represents the area of intersection of ground truth and detection bounding boxes, $GT\cup DT$ represents the area of union of ground truth and detection bounding boxes. However, when there is misalignment between modalities, the coordinates of each object in both modalities are not the same. If we only concern about precision of one modality, another modality will have poor precision. In order to measure the ability to handle with both modalities, especially when level of misalignment is high, we introduce a new evaluation metric, which we call ``multi-modal IoU (IoU\textsuperscript{M})" defined as:
\begin{align}
IoU\textsuperscript{M} = \frac{(GT^V\cap DT^V)+(GT^T\cap DT^T)}{(GT^V\cup DT^V)+(GT^T\cup DT^T)},
\end{align}
where $GT^V$, $GT^T$ denote paired ground truth bounding boxes referring to the same object from visible and thermal modality respectively. $DT^V$, $DT^T$ denote paired detection bounding boxes referring to the same object from visible and thermal modality respectively. IoU\textsuperscript{M} can be used to determine the precision of detection bounding boxes in both modalities. Moreover, in order to thoroughly evaluate each modality, we define visible IoU (IoU\textsuperscript{V}) as IoU in visible modality, and thermal IoU (IoU\textsuperscript{T}) as IoU in thermal modality.

\vspace{0.15cm}
\noindent
\textbf{Mini-batch sampling.}
We follow sampling strategies from~\cite{faster} and~\cite{fast}. But since our approach has one regressor for each modality exclusively and we need to keep paired relations for all proposals and RoIs, we select training samples as anchor pairs and RoI pairs. For this purpose, we use IoU\textsuperscript{M} as selection criteria instead of IoU. For RPN, we assign positive and negative labels to anchor pairs that have IoU\textsuperscript{M} overlap higher than 0.63 with any ground truth bounding box pair and lower than 0.3 for all ground truth bounding box pairs respectively. For detector, we assign positive and negative labels to RoI pairs that have IoU\textsuperscript{M} overlap with any ground truth bounding box pair higher than 0.5 and lower than 0.5 but higher than 0.1 respectively.

\begin{table}[t]
  \caption{{\bf{Comparison with state-of-the-art methods on KAIST dataset, with simulated disparity of misalignment, by MR\textsuperscript{M}.}}}
  \vspace{-6mm}
  \begin{center}
  \scalebox{0.59}{
    \begin{tabular}{| c | c c c c | c c c c |}
      \hline
      \multirow{4}{*}{\shortstack{Thermal\\Shift\\Distance}} & \multicolumn{8}{c|}{MR\textsuperscript{M}}\\
      \cline{2-9}
      & \multicolumn{4}{c|}{IoU\textsuperscript{M} threshold: 0.5} & \multicolumn{4}{c|}{IoU\textsuperscript{M} threshold: 0.7}\\
      \cline{2-9}
      & MSDS & AR-CNN & MBNet & Ours & MSDS & AR-CNN & MBNet & Ours\\
      & \cite{msds} & \cite{arcnn} & \cite{mbnet} & & \cite{msds} & \cite{arcnn} & \cite{mbnet} & \\
      \hline
      -20 & 84.15 & 82.05 & 82.13 & \textbf{62.24} & 98.27 & 98.12 & 97.06 & \textbf{76.46}\\
      -15 & 59.67 & 59.80 & 58.19 & \textbf{37.51} & 93.41 & 94.88 & 92.65 & \textbf{62.12}\\
      -10 & 27.06 & 21.61 & 23.13 & \textbf{16.71} & 85.06 & 89.47 & 82.98 & \textbf{49.52}\\
      -5 & 13.77 & \textbf{9.65} & 10.01 & 11.15 & 63.15 & 66.04 & 55.96 & \textbf{41.66}\\
      %-3 & 12.32 & 8.32 & \textbf{8.21} & 9.92 & 55.41 & 54.33 & 46.02 & \textbf{40.81}\\
      0 & 11.09 & 8.79 & \textbf{7.76} & 9.67 & 51.50 & 42.14 & 38.95 & \textbf{38.65}\\
      %3 & 12.27 & \textbf{9.46} & \textbf{9.46} & 10.09 & 55.71 & 47.91 & 44.66 & \textbf{37.76}\\
      5 & 13.73 & \textbf{10.35} & \textbf{10.35} & 11.14 & 62.10 & 56.75 & 55.41 & \textbf{39.55}\\
      10 & 27.48 & 19.84 & 21.51 & \textbf{17.10} & 87.49 & 87.22 & 84.52 & \textbf{45.46}\\
      15 & 60.08 & 52.85 & 55.49 & \textbf{35.44} & 96.37 & 96.43 & 95.10 & \textbf{58.45}\\
      20 & 86.23 & 82.10 & 84.15 & \textbf{59.93} & 98.88 & 99.14 & 98.55 & \textbf{74.35}\\
      \hline
    \end{tabular}
  }        
  \end{center}
\vspace{-6mm}
\end{table}

\section{Experiments}
Detection performance was measured by log-average miss rate (MR) suggested by~\cite{toolbox}. MR is defined by geometrical mean of miss rates at nine specific false positives per image (FPPI), at which we divide them by evenly spaced FPPI in range of [10\textsuperscript{-2}, 10\textsuperscript{0}]. Since number of false negatives and false positives are required to calculate miss rate and FPPI, IoU is used to determine objects and detection results into those categories with threshold of 0.5 and 0.7. In order to evaluate precision of detection results in both modalities, visible MR (MR\textsuperscript{V}) representing MR based on IoU\textsuperscript{V}, thermal MR (MR\textsuperscript{T}) representing MR based on IoU\textsuperscript{T}, and multi-modal MR (MR\textsuperscript{M}) representing MR based on IoU\textsuperscript{M}, were used in our experiments. Furthermore, to evaluate the effectiveness of our method against misalignment, we simulated disparity of misalignment between modalities by shifting thermal images horizontally from 0 to 20 pixels in both directions. All experiments were performed under reasonable configuration~\cite{kaist}, i.e., only pedestrians taller than 55 pixel under partial or no occlusion are considered. For other methods that do not have both DT\textsuperscript{V} and DT\textsuperscript{T}, we substituted both with their detection bounding boxes.

\vspace{0.15cm}
\noindent
{\bf{Dataset.}}
We used KAIST dataset~\cite{kaist} in our experiments. It was recorded in both day and night to consider changes in light conditions.
%Most of image pairs are well aligned.
Since we focus on misalignment, we adopted annotations provided by L. Zhang et al.~\cite{arcnn}, which localize objects in each modality independently and keep all of their paired relations, as ground truth. Only 2,252 frames from test set were used in performance test as traditional.

\vspace{0.15cm}
\noindent
{\bf{Implementation details.}}
We adopt VGG-16~\cite{vgg} pre-trained on ImageNet~\cite{imagenet,imagenet_classification} as our two-stream backbone networks as in AR-CNN~\cite{arcnn}. 
%for fair comparison. 
We train the network for 3 epochs with learning rate of 0.005 and 1 additional epoch with learning rate of 0.0005 by Stochastic Gradient Descent (SGD) optimizer with 0.9 momentum and 0.0005 weight decay. We select 8,892 images %from training set
containing informative pedestrians
for the training. Image resolution is fixed to 640×512.
%All images are horizontally flipped in the training for data augmentation.
All images are horizontally flipped for data augmentation.

\begin{table}[t]
  \caption{{\bf{Comparison with state-of-the-art methods on KAIST dataset, without simulated disparity of misalignment, by MR\textsuperscript{V}, MR\textsuperscript{T}, MR\textsuperscript{M}.}}}
  \vspace{-6mm}
  \begin{center}
      \scalebox{0.75}{
        \begin{tabular}{| c | c | c c c c |}
          \hline
          \multicolumn{2}{|c|}{\multirow{2}{*}{Method}} & MSDS & AR-CNN & MBNet & Ours\\
          \multicolumn{2}{|c|}{} & \cite{msds} & \cite{arcnn} & \cite{mbnet} & \\
          \hline
          \multirow{2}{*}{MR\textsuperscript{V}} & IoU\textsuperscript{V} threshold: 0.5 & 11.28 & 9.86  & \textbf{7.89} & 10.69\\
          & IoU\textsuperscript{V} threshold: 0.7                       & 50.01 & 43.50 & 41.90         & \textbf{41.45}\\
          \hline
          \multirow{2}{*}{MR\textsuperscript{T}} & IoU\textsuperscript{T} threshold: 0.5 & 12.51 & 8.26  & \textbf{8.12} & 9.24\\
          & IoU\textsuperscript{T} threshold: 0.7                       & 55.56 & 43.55 & 41.27         & \textbf{39.14}\\
          \hline
          \multirow{2}{*}{MR\textsuperscript{M}} & IoU\textsuperscript{M} threshold: 0.5 & 11.09 & 8.79  & \textbf{7.76} & 9.67\\
          & IoU\textsuperscript{M} threshold: 0.7                       & 51.50 & 42.14 & 38.95         & \textbf{38.65}\\
          \hline
        \end{tabular}
      }
  \end{center}
\vspace{-6mm}
\end{table}

\vspace{0.15cm}
\noindent
{\bf{Comparison with state-of-the-art methods.}}
 We selected three state-of-the-art methods for our experiments, MSDS (MSDS-RCNN)~\cite{msds} is representative for methods without misalignment consideration, AR-CNN~\cite{arcnn} and MBNet~\cite{mbnet} are methods that consider misalignment. 
 % None of them infers location of each object in both modalities separately beside us. 
 From Table 1, when IoU\textsuperscript{M} threshold is 0.5, we only achieve the lowest MR when the misalignment is larger than 10 pixels. However, when IoU\textsuperscript{M} threshold is 0.7, we achieve the lowest MR at all shift distances, which indicates our proposed method's robustness to large misalignment.

From Table 2, our method's performance is comparable to state-of-the-art methods. Still, it is not the best on any evaluation metrics when IoU thresholds are 0.5. However, when IoU thresholds increase to 0.7, i.e., requirement of precision is higher, our method achieves the best performance among available competitors in all evaluation metrics, which demonstrates our superior precision of detection bounding boxes in both modalities. This can benefit applications that require high and reliable precision of detection, such as autonomous vehicle, precise location of pedestrians is crucial.

% \subsection{Ablation study} (Anticipate)
% To verify the effectiveness of our approach, we conduct experiments with a detail analysis of our model. all detectors are trained with the same setting. From Table 3, we can see that multi-modal RPN and double regressor detector both help reducing MR and significantly bring down MR\textsuperscript{M} especially when the shift distance is high, which demonstrate their effectiveness against large misalignment. 
  
% \begin{table}[t]
% \caption{Detection results .}
%   \begin{center}
%   \scalebox{0.70}{
%     \begin{tabular}{| c | c | c | c | c |}
%       \hline
%       \begin{tabular}{@{}c@{}c@{}}Thermal\\Shift\\Distance\end{tabular} &\begin{tabular}{@{}c@{}c@{}}Multi-\\Modal\\RPN\\\end{tabular} &  \begin{tabular}{@{}c@{}c@{}}Multi-\\ Modal\\Detector\end{tabular}  &  \makebox[9mm]{MR} &  \makebox[9mm]{MR\textsuperscript{M}}\\
%       \hline
%       & & & 0 & 0 \\
%       0 & \ding{51} & & 0 & 0 \\
%       & & \ding{51} & 0 & 0 \\
%       & \ding{51} & \ding{51} & 0 & 0 \\
%       \hline
%       & & & 0 & 0 \\
%       5 & \ding{51} & & 0 & 0 \\
%       & & \ding{51} & 0 & 0 \\
%       & \ding{51} & \ding{51} & 0 & 0 \\
%       \hline
%       & & & 0 & 0 \\
%       10 & \ding{51} & & 0 & 0 \\
%       & & \ding{51} & 0 & 0 \\
%       & \ding{51} & \ding{51} & 0 & 0 \\
%       \hline
%     \end{tabular}
%   }
%     \label{sample-table}
%   \end{center}
% \vspace{-6mm}
% \end{table}

\section{Conclusion}
% In this paper, we have analyzed the current misalignment problem of existing methods. 
We have proposed the novel multi-modal detection method based on modal-wise regression and multi-modal IoU, the proposed method is robust to large misalignment and also keeps paired relations of all detection bounding boxes between both modalities. To the best of our knowledge, this paper is the first work to tackle the problem: detection from multi-modal images with large misalignment. 
Our experiments showed that when the precision requirement of the bounding box or the level of misalignment is high, our proposed method achieves the best performance, demonstrating our robustness to misalignment and superior precision of detection bounding boxes in both modalities. 
% However, there is no method that infers position of objects for both modalities independently beside ours currently, so it is not totally fair to compare by MR\textsuperscript{M} yet. We hope that future research will consider our concern and use MR\textsuperscript{M} to evaluate multi-modal detection with misalignment.    

\balance
% \nocite{*}
\bibliographystyle{plain}
\bibliography{ref}

\begin{thebibliography}{10}

\bibitem{kaistdriving}
Yukyung {Choi}, Namil {Kim}, Soonmin {Hwang}, Kibaek {Park}, Jae~Shin {Yoon},
  Kyounghwan {An}, and In~So {Kweon}.
\newblock Kaist multi-spectral day/night data set for autonomous and assisted
  driving.
\newblock {\em IEEE Transactions on Intelligent Transportation Systems},
  19(3):934--948, 2018.

\bibitem{imagenet}
Jia {Deng}, Wei {Dong}, Richard {Socher}, Li-Jia {Li}, {Kai Li}, and {Li
  Fei-Fei}.
\newblock Imagenet: A large-scale hierarchical image database.
\newblock In {\em IEEE Conference on Computer Vision and Pattern Recognition
  (CVPR)}, pages 248--255, 2009.

\bibitem{toolbox}
Piotr {Dollar}, Christian {Wojek}, Bernt {Schiele}, and Pietro {Perona}.
\newblock Pedestrian detection: An evaluation of the state of the art.
\newblock {\em IEEE Transactions on Pattern Analysis and Machine Intelligence
  (PAMI)}, 34(4):743--761, 2012.

\bibitem{acf}
Piotr {Dollár}, Ron {Appel}, Serge {Belongie}, and Pietro {Perona}.
\newblock Fast feature pyramids for object detection.
\newblock {\em IEEE Transactions on Pattern Analysis and Machine Intelligence
  (PAMI)}, 36(8):1532--1545, 2014.

\bibitem{dong2018learning}
Jing Dong, Byron Boots, Frank Dellaert, Ranveer Chandra, and Sudipta Sinha.
\newblock Learning to align images using weak geometric supervision.
\newblock In {\em International Conference on 3D Vision (3DV)}, pages 700--709.
  IEEE, 2018.

\bibitem{fast}
Ross {Girshick}.
\newblock Fast r-cnn.
\newblock In {\em Proceedings of the IEEE International Conference on Computer
  Vision (ICCV)}, pages 1440--1448, 2015.

\bibitem{pedestriandaynight}
Alejandro González, Zhijie Fang, Yainuvis Socarras, Joan Serrat, David
  Vázquez, Jiaolong Xu, and Antonio~Manuel López.
\newblock Pedestrian detection at day/night time with visible and fir cameras:
  A comparison.
\newblock {\em Sensors}, 16(6), 2016.

\bibitem{fusionseg}
Dayan Guan, Yanpeng Cao, Jiangxin Yang, Yanlong Cao, and Michael~Ying Yang.
\newblock Fusion of multispectral data through illumination-aware deep neural
  networks for pedestrian detection.
\newblock {\em Information Fusion}, 50:148--157, 2019.

\bibitem{accum}
{Hangil Choi}, Seungryong {Kim}, {Kihong Park}, and Kwanghoon {Sohn}.
\newblock Multi-spectral pedestrian detection based on accumulated object
  proposal with fully convolutional networks.
\newblock In {\em 2016 23rd International Conference on Pattern Recognition
  (ICPR)}, pages 621--626, 2016.

\bibitem{maskrcnn}
Kaiming {He}, Georgia {Gkioxari}, Piotr {Dollár}, and Ross {Girshick}.
\newblock Mask r-cnn.
\newblock In {\em Proceedings of the IEEE International Conference on Computer
  Vision (ICCV)}, pages 2980--2988, 2017.

\bibitem{kaist}
Soonmin {Hwang}, Jaesik {Park}, Namil {Kim}, Yukyung {Choi}, and In~So {Kweon}.
\newblock Multispectral pedestrian detection: Benchmark dataset and baseline.
\newblock In {\em IEEE Conference on Computer Vision and Pattern Recognition
  (CVPR)}, pages 1037--1045, 2015.

\bibitem{liu}
Shu~Wang Jingjing~Liu, Shaoting~Zhang and Dimitris Metaxas.
\newblock Multispectral deep neural networks for pedestrian detection.
\newblock In {\em British Machine Vision Conference (BMVC)}, pages 73.1--73.13,
  2016.

\bibitem{kim2015dasc}
Seungryong Kim, Dongbo Min, Bumsub Ham, Seungchul Ryu, Minh~N Do, and Kwanghoon
  Sohn.
\newblock Dasc: Dense adaptive self-correlation descriptor for multi-modal and
  multi-spectral correspondence.
\newblock In {\em IEEE Conference on Computer Vision and Pattern Recognition
  (CVPR)}, pages 2103--2112, 2015.

\bibitem{imagenet_classification}
Alex Krizhevsky, Ilya Sutskever, and Geoffrey~Everest Hinton.
\newblock Imagenet classification with deep convolutional neural networks.
\newblock In {\em Advances in Neural Information Processing Systems}, pages
  1097--1105. 2012.

\bibitem{rpnmulti}
Daniel {König}, Michael {Adam}, Christian {Jarvers}, Georg {Layher}, Heiko
  {Neumann}, and Michael {Teutsch}.
\newblock Fully convolutional region proposal networks for multispectral person
  detection.
\newblock In {\em IEEE Conference on Computer Vision and Pattern Recognition
  Workshops (CVPRW)}, pages 243--250, 2017.

\bibitem{msds}
Chengyang Li, Dan Song, Ruofeng Tong, and Min Tang.
\newblock Multispectral pedestrian detection via simultaneous detection and
  segmentation.
\newblock In {\em British Machine Vision Conference (BMVC)}, 2018.

\bibitem{liu2019illumination}
Chengyang Li, Dan Song, Ruofeng Tong, and Min Tang.
\newblock Illumination-aware faster r-cnn for robust multispectral pedestrian
  detection.
\newblock {\em Pattern Recognition}, 85:161--171, 2019.

\bibitem{scaleaware}
Jianan {Li}, Xiaodan {Liang}, Shengmei {Shen}, Tingfa {Xu}, Jiashi {Feng}, and
  Shuicheng {Yan}.
\newblock Scale-aware fast r-cnn for pedestrian detection.
\newblock {\em IEEE Transactions on Multimedia}, 20(4):985--996, 2018.

\bibitem{li2013image}
Shutao Li, Xudong Kang, and Jianwen Hu.
\newblock Image fusion with guided filtering.
\newblock {\em IEEE Transactions on Image processing (TIP)}, 22(7):2864--2875,
  2013.

\bibitem{whathelp}
Jiayuan {Mao}, Tete {Xiao}, Yuning {Jiang}, and Zhimin {Cao}.
\newblock What can help pedestrian detection?
\newblock In {\em IEEE Conference on Computer Vision and Pattern Recognition
  (CVPR)}, pages 6034--6043, 2017.

\bibitem{ogino2017coaxial}
Yuka Ogino, Takashi Shibata, Masayuki Tanaka, and Masatoshi Okutomi.
\newblock Coaxial visible and fir camera system with accurate geometric
  calibration.
\newblock In {\em Thermosense: Thermal Infrared Applications XXXIX}, volume
  10214, page 1021415. International Society for Optics and Photonics, 2017.

\bibitem{unified}
Kihong Park, Seungryong Kim, and Kwanghoon Sohn.
\newblock Unified multi-spectral pedestrian detection based on probabilistic
  fusion networks.
\newblock {\em Pattern Recognition}, 80:143--155, 2018.

\bibitem{faster}
Shaoqing Ren, Kaiming He, Ross Girshick, and Jian Sun.
\newblock Faster r-cnn: Towards real-time object detection with region proposal
  networks.
\newblock In {\em Advances in Neural Information Processing Systems}, pages
  91--99, 2015.

\bibitem{10.1117/1.JEI.28.3.033028}
Thapanapong Rukkanchanunt, Masayuki Tanaka, and Masatoshi Okutomi.
\newblock {Full thermal panorama from a long wavelength infrared and visible
  camera system}.
\newblock {\em Journal of Electronic Imaging}, 28(3):1 -- 10, 2019.

\bibitem{shibata2017misalignment}
Takashi Shibata, Masayuki Tanaka, and Masatoshi Okutomi.
\newblock Misalignment-robust joint filter for cross-modal image pairs.
\newblock In {\em Proceedings of the IEEE International Conference on Computer
  Vision (ICCV)}, pages 3295--3304, 2017.

\bibitem{vgg}
Karen Simonyan and Andrew Zisserman.
\newblock Very deep convolutional networks for large-scale image recognition.
\newblock In {\em International Conference on Learning Representations (ICLR),
  San Diego, CA, USA, May 7-9, 2015, Conference Track Proceedings}, 2015.

\bibitem{cats}
Wayne Treible, Philip Saponaro, Scott Sorensen, Abhishek Kolagunda, Michael
  O'Neal, Brian Phelan, Kelly Sherbondy, and Chandra Kambhamettu.
\newblock Cats: A color and thermal stereo benchmark.
\newblock In {\em IEEE Conference on Computer Vision and Pattern Recognition
  (CVPR)}, 2017.

\bibitem{Wagner2016MultispectralPD}
J{\"o}rg Wagner, Volker Fischer, Michael Herman, and Sven Behnke.
\newblock Multispectral pedestrian detection using deep fusion convolutional
  neural networks.
\newblock In {\em European Symposium on Artificial Neural Networks,
  Computational Intelligence and Machine Learning (ESANN)}, 2016.

\bibitem{learningcrossmodal}
Dan {Xu}, Wanli {Ouyang}, Elisa {Ricci}, Xiaogang {Wang}, and Nicu {Sebe}.
\newblock Learning cross-modal deep representations for robust pedestrian
  detection.
\newblock In {\em IEEE Conference on Computer Vision and Pattern Recognition
  (CVPR)}, pages 4236--4244, 2017.

\bibitem{ccf}
Bin Yang, Junjie Yan, Zhen Lei, and Stan~Z. Li.
\newblock Convolutional channel features.
\newblock In {\em Proceedings of the IEEE International Conference on Computer
  Vision (ICCV)}, 2015.

\bibitem{guide}
Heng Zhang, Elisa Fromont, Sebastien Lefevre, and Bruno Avignon.
\newblock Guided attentive feature fusion for multispectral pedestrian
  detection.
\newblock In {\em Proceedings of the IEEE/CVF Winter Conference on Applications
  of Computer Vision (WACV)}, pages 72--80, 2021.

\bibitem{fasterforpedestrian}
Liliang Zhang, Liang Lin, Xiaodan Liang, and Kaiming He.
\newblock Is faster r-cnn doing well for pedestrian detection?
\newblock In {\em Proceedings of the European Conference on Computer Vision
  (ECCV)}, pages 443--457, 2016.

\bibitem{zhang2019cross}
Lu~Zhang, Zhiyong Liu, Shifeng Zhang, Xu~Yang, Hong Qiao, Kaizhu Huang, and
  Amir Hussain.
\newblock Cross-modality interactive attention network for multispectral
  pedestrian detection.
\newblock {\em Information Fusion}, 50:20--29, 2019.

\bibitem{arcnn}
Lu~Zhang, Xiangyu Zhu, Xiangyu Chen, Xu~Yang, Zhen Lei, and Zhiyong Liu.
\newblock Weakly aligned cross-modal learning for multispectral pedestrian
  detection.
\newblock In {\em Proceedings of the IEEE International Conference on Computer
  Vision (ICCV)}, 2019.

\bibitem{occlusionaware}
Shifeng Zhang, Longyin Wen, Xiao Bian, Zhen Lei, and Stan~Z. Li.
\newblock Occlusion-aware r-cnn: Detecting pedestrians in a crowd.
\newblock In {\em Proceedings of the European Conference on Computer Vision
  (ECCV)}, 2018.

\bibitem{mbnet}
Kailai Zhou, Linsen Chen, and Xun Cao.
\newblock Improving multispectral pedestrian detection by addressing modality
  imbalance problems.
\newblock In {\em Proceedings of the European Conference on Computer Vision
  (ECCV)}, pages 787--803, 2020.

\end{thebibliography}

\end{document}